# Deep Learning for Vision-Based Fall Detection System: Enhanced Optical Dynamic Flow

**Sagar Chhetri[1], Abeer Alsadoon[1*], Thair Al-Dala'in[1,2], P.W.C. Prasad[1], Tarik A. Rashid[3], Angelika Maag[1]**

[1]School of Computing and Mathematics, Charles Sturt University, Sydney Campus, Australia
[2]School of computing engineering and mathematics, Western Sydney University, Sydney, Australia
[3]Computer Science and Engineering, University of Kurdistan Hewler, Erbil, KRG, Iraq.

*Abeer Alsadoon[1*]*

* Corresponding author. Dr. Abeer Alsadoon, [1]School of Computing and Mathematics, Charles Sturt University, Sydney Campus, Australia, Email: alsadoon.abeer@gmail.com, Phone +61 2 9291 9387

## Abstract

Accurate fall detection for the assistance of older people is crucial to reduce incidents of deaths or injuries due to falls. Meanwhile, a vision-based fall detection system has shown some significant results to detect falls. Still, numerous challenges need to be resolved. The impact of deep learning has changed the landscape of the vision-based system, such as action recognition. The deep learning technique has not been successfully implemented in vision-based fall detection systems due to the requirement of a large amount of computation power and the requirement of a large amount of sample training data. This research aims to propose a vision-based fall detection system that improves the accuracy of fall detection in some complex environments such as the change of light condition in the room. Also, this research aims to increase the performance of the pre-processing of video images. The proposed system consists of the Enhanced Dynamic Optical Flow technique that encodes the temporal data of optical flow videos by the method of rank pooling, which thereby improves the processing time of fall detection and improves the classification accuracy in dynamic lighting conditions. The experimental results showed that the classification accuracy of the fall detection improved by around 3% and the processing time by 40 to 50ms. The proposed system concentrates on decreasing the processing time of fall detection and improving classification accuracy. Meanwhile, it provides a mechanism for summarizing a video into a single image by using a dynamic optical flow technique, which helps to increase the performance of image pre-processing steps.

## 1. Introduction

The traditional techniques used for fall detection system is based on calculating different features of the human body, which is under surveillance. These systems consist of human shape analysis through extracting the shape features and using them for the classification purpose and posture as fall. However, the limitation of conventional approaches has required background and foreground detection process, which is representing the challenge to implementing it in a complicated situation such as occlusion and variation in the postures [1]. Furthermore, the traditional approaches have difficulty in recognizing a regular fall from other daily life activities (ALD) such as, lying down, sitting, crouching, which are similar to regular falls.

In the field of machine learning and computer vision, deep learning techniques, such as convolution neural networks (CNN) achieved high performance for image classification and motion recognition. CNN enables the automatic extraction of hierarchical image features from a large amount of annotated data by layer-wise pre-training and fine-tuning [1]. Wang et al. [2] proposed an approach that is used as a PCNet deep network that extracts the features from RGB images and then used SVM as a classifier to detect fallen. However, using deep learning for image and action recognition requires a lot of computational power and a large amount of trained data [3]. Thus, it is recommended by many researchers to use a pre-trained model, such as ImageNet. This type of model can contain 14 million images of various categories to extract generic patterns and then fine-tune the network to extract features that are related to a particular domain. Various techniques of deep learning are used for fall detection. However, there is a limited number of datasets used for fall detection, so such a pre-trained model needs to be considered to enhance current systems.

The deep learning architectures have shown excellent performance accuracy in terms of sensitivity and processing time. Deep learning shows significant results in detecting a regular fall from other daily life activities. Núñez-Marcos et al. [3] used VGG16 Net as a type of deep convolutional network architecture and optical flow for image processing. They achieved an accuracy of 83.02%. The proposed approach improved its state-of-the-art model [2] in various standard public datasets. However, Wang et al. [2] used the optical flow method for image pre-processing, which is required a large amount of computation time, and light in the video to be static. In the real-world situation, the lighting condition may be dynamic, which may be resulted in an unexpected displacement field from optical flow and produces low classification accuracy.

The purpose of this paper is to increase the classification accuracy in dynamic lighting conditions and decrease the pre-processing time by using dynamic optical flow techniques. However, using optical flow requires sizeable computational power. Besides, using optical flow techniques in dynamic lighting conditions resulted in displacement vectors, which is not desirable and produced a poor classification accuracy. Therefore, in this paper, a rank pooling technique has been used to summarize the optical flow frames into a single image. This technique is called a dynamic optical image that can capture every dynamic action in the video. The technique will reduce pre-processing time and increase the classification accuracy of the system.

The rest sections of the paper are organized as follows: Section 2 presents a literature review. Section3 presents the state of the art and the proposed method. Section 4 presents the results and discussion of the proposed method. Section 5 presents the study conclusion.

## 2. Literature Review

The literature review aims to review the state of art and provides a perception of different methods, techniques, tools that have been applied in the area of this study. The literature review is divided into two sections. The first section explores the existing methods and models used in feature extraction. The second section explores the methods used in fall detection.

*Feature extraction:*

Fan et al. [1] investigated the performance of multidirectional statistical analysis, in terms of classification accuracy of the human posture. The statistical analysis was done by using a normalized directional histogram around the center of the extracted subject's ellipse to represents a human posture. This process led to an accuracy of 97.1% for recognizing four human postures (sitting, crouching down, and walking). This research was limited to do recognition for just four basic human postures. In a real-world situation, a fall accident involves additional complicated human posture movement than only four basic human posture. Hnoohom et al. [4] proposed a method for automating feature extraction using a deep CNN and global maximum pooling technique for enhancing model discriminability. The results showed that the proposed method achieved better performance and higher accuracy compared with other methods such as dynamic time warping recurrent neural networks, and multi-layer perceptron techniques. Zerrouki et al. [5] investigated the effectiveness of the Adaptive Boosting Algorithm, in terms of classification accuracy of the human activity. The proposed system was done by extracting features of the human body through background subtraction and human silhouette, which is divided into five segments based on the body's center of gravity. The results led to an accuracy of 93.91 % of human action classification and consumed 680 ms for average training and 87ms for testing. The proposed prototype leads to errors and false in the classification when extracting the human body from an automatic changing background. Wang et al. [2] investigated the performance of the SVM classifier, in terms of sensitivity and specificity for fall detection. They extracted the features of persona falling images by PCANet deep network. The results achieved 89.2% sensitivity and 90.3% accuracy of the SVM classifier. The achieved results have overcome the limitation of its state-of-the-art techniques. However, the data sample which has been used is considered not sufficient. Therefore, the accuracy of the proposed method will fall if a large number of videos have been utilized. As a result, the proposed prototype does not provide any further improvement for current fall detection systems. Saleh et al. [6] proposed a low computational cost fall detection algorithm using machine learning-based. They used a novel online feature extraction method that employs the time characteristics of falls. Also, a novel design of a machine learning-based system is proposed to achieve the best accuracy/numerical complexity tradeoff. Experimental results showed that the accuracy of the proposed algorithm was 99.9% with a computational cost of less than 500 floating-point operations per second.

*Fall detection:*

Ali et al. [7] investigated the effectiveness of J48 and AdaBoost classifiers, in terms of classification accuracy and execution time for fall detection. The proposed fall prototype was implemented by using the variance of various discriminatory features like motion, geometric orientation, and geometric location. The results led to 99.03% fall detection accuracy. The execution time was 0.01s for the J48 classifier and 0.025s for the AdaBoost classifier. However, the performance of those classifiers reduces a complicated situation, such as the same color of clothes and background, multisource, and multiple people with occlusion. Xiong [8] proposed a skeleton-based 3D consecutive-low-pooling neural network (S3D-CNN) for fall detection. The proposed system was evaluated on public and self-collected datasets and achieved the best results compared to existing methods. Min et al. [9] investigated the performance of faster RCNN, in terms of the area under the ROC curve. The proposed system used scene analysis for fall detection. They proposed a new method for human fall detection on furniture based on deep learning and activity characteristics. Also, include other human characteristics: motion speed, centroid, and human aspect ratio. The result led to 0.941 AUC and 93% accuracy.

In contrast, the proposed approach only takes the spatial relation of the human and furniture for fall detection. This approach did not consider the temporal extent of fall, which is deemed to be necessary as the features of the spatial relations. By including the temporal extent with the spatial relation features, it might provide additional correct results and improve the system performance. Zhang et al. [10] proposed a trajectory-weighted deep convolutional rank-pooling descriptor for fall detection in videos. Zerrouki et al. [11] investigated the effectiveness of a hidden Markov model to classify human posture and discover fall, in terms of classification accuracy. The proposed system implemented by using Support Vector Machine (SVM) for posture classification and HMM model to detect human activities. The result led to 95.26% accuracy of the human activities and 96.88% accuracy in detecting falls using SVM-HMM. The proposed intended methodology is used as a background subtraction method to extract the human silhouette using RGB cameras that are not capable of identifying human silhouettes in a dark environment situation. Therefore, it is not appropriate for fall detection systems. Wang et al. [12] proposed a multi-sensor-based fall detection system. They presented a Multisource CNN Ensemble (MCNNE) structure to improve the detection accuracy. They found that MCNNE has a better performance compared to a single CNN structure and various ensemble bi-model structures. Cheng et al. [13] used a cascade AdaBoost SVM classifier in the tri-axial accelerometer-based fall detection. Hnoohom et al. [14] used accelerometer and gyroscope sensor data to compare the performance of traditional ensemble learning. The study results showed that ensemble learning-based approaches could improve the detection accuracy whether the sensor is placed on the arm or the waist.

## 3. Materials and Methods

### 3.1 State of the art

Fig. 1 represents the features of the current system. The useful features are highlighted inside the broken blue line. The limitations in the system are highlighted inside the broken red border. The system was proposed by Núñez-Marcos et al. [3]. They used deep CNN to decide if a video contains a person falling or not. This approach uses optical flow images as an input to the deep network. However, the optical flow images ignore any appearance-related features such as color, contrast, and brightness. The proposed approach minimizes the hand-crafted image processing steps by using CNN. CNN can learn a set of features and improved the performance when enough examples are provided during the training phase. However, the proposed system has been made more generic.

Núñez-Marcos et al. [3] presented a vision-based fall detection system using a CNN, which applies transfer learning from the action recognition domain to fall detection. Three different public datasets were used to evaluate the proposed approach. This model consists of two main stages, as shown in Fig. 1: Pre-processing stage, and feature extraction, and classification stage.

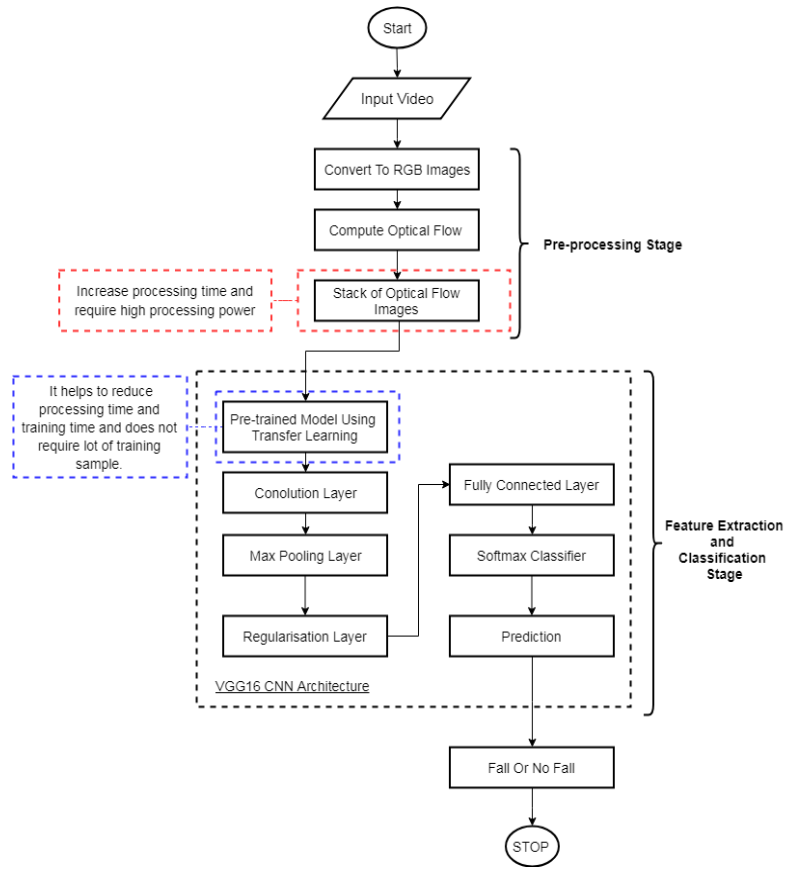

**Figure 1**: Block diagram of the state-of-the-art system [3]

[The blue borders show the good features of the solution, and the red border show the limitation of it]

*Pre-processing stage:*

The video consists of contiguous frames stacked together. For a fall detection system, just a few adjacent frames are needed to detect a fall in the video. Besides that, some techniques process each frame individually and do not consider the correlation between the stacked frames. To solve this problem, Núñez-Marcos et al. [3] utilized an optical flow algorithm that describes the displacement vector between two frames. The optical flow generator receives consecutive images and uses the TVL-1 optical flow algorithm [15][16] that considers just motion in the video and removes any static background, as shown in Table 1. Although static background can be removed from the motion of video, the perfect foreground cannot be obtained by utilizing the optical flow technique alone due to various background noise, which may arise from the change in brightness. Background noise can emerge randomly because the optical flow technique analyzes the whole image and computes the flow for every pixel. Due to the presence of background noise, the accuracy of flow estimation decreased in the low light situation. Additionally, computing flow estimation for every pixel and generating a stack of optical flow images requires very high processing power, which is eventually increased the estimating of processing time. New solutions try to overcome general optical flow problem problems such as constant brightness assumption. This assumption is based on three successive images of a sequence. The approach assumes that motion is translational on a region large enough to regularize the aperture problem. In the way of avoiding outliers, a robust multi-resolution method was used. It is composed of a low-pass pyramid and an M-estimator technique. This method offered some good results on artificial and natural image sequences [17].

The optical flow algorithm represents the patterns of the motion of objects as displacement vector fields between two consecutive images. Núñez-Marcos et al. [3] selected the TVL-1 optical flow algorithm due to its performance with changing lighting conditions compared to another algorithm. However, the TVL-1 optical flow algorithm can capture just small events. In a real-world situation, the human fall event involves many complex human actions. TVL-1 algorithm does not require modeling the dynamics of video content, which is crucial to increase the accuracy of any human activities such as fall. Eq. (1) describes the TVL-1 optical flow equation [18] [3] :

$$E = \int_{\Omega} \{\lambda |I_0(x) - I_1(x + u(x))| + |\nabla u|\} dx \qquad (1)$$

Where,

$I_0$ and $I_1$ is the image pair,
x the point of a pixel in the x-axis
$u = (u_1(x), u_2(x))^T$ is the two-dimension displacement field
$\lambda$ is a relative weighting term.
E is the energy function

TVL-1 has not been considered a temporal evolution of human action. The first part of the TVL-1 of Eq. (1) is the data term that includes a brightness consistency assumption, which measures the accuracy with velocity field that describes the observable image motion. The second part is the regularization term, which penalizes high variations in the optical flow field to obtain smooth displacement fields. The regularization term in TVL-1 optical flow poses several issues. Some of them are not feasible in the learning of 'flow patterns' if different image structures are given, and the motion of the camera may give rise to a multitude of motion patterns with little resemblance between motion fields from different videos [19]. So, it is necessary to use an appropriate regularization term. Due to these issues with the regularization term, it is difficult to obtain accurate optical flow estimation.

*Feature extraction and classification stage*

In this stage, CNN was used for feature extraction and classification of images. In particular, VGG-16 CNN was used. The architecture of VGG-16 was chosen because of its high accuracy in the feature extraction process. VGG-16 CNN architecture consists of 3 layers: convolution layers, max-pooling layers, and pooling layers. In convolution layers, VGG-16 restricts the use of 3x3 convolution kernels size. The choice of small kernel size leads to reduce the number of parameters that can result in practical training and testing.

Most importantly, with a small kernel size, it can stack more layers in deep networks; as a result, it would increase the performance of fall detection. By stacking series of 3x3 size kernel filters, the effective receptive can be increased to larger values, for example, 5x5 with two layers, 7x7 size with three layers [20]. In VGG architecture, each convolution layer was followed by the rectified linear unit (ReLU) layer. Pooling layers operate on the blocks of the input feature map and combines the featured activations. This combination task is defined by a max function, which selects the maximum activation from the chosen group of blocks. The Max-pooling layer minimizes the spatial size of the feature, and it helps in reducing the number of parameters and the amount of computation in the network [20]. A fully connected layer takes the input from the ReLU layer and generates the class score that is used for the classification of fall detection. VGG16 architecture uses dropout layers in the first two fully connected layers to avoid over-fitting. The input layer of VGG-16 accepts a stack of optical flow images. The ImageNet dataset was utilized to address a small number of fall samples in the dataset. Besides, it used to learn the generic features as it has 14 million images. Based on the CNN-trained ImageNet dataset, the input to the CNN is modified to accept input images of size 224x224 and stack size of 20. The network was retrained on the optical flow stack of the UCF101 dataset. In the final step of transfer learning, the weight of the convolution layer was frozen to make the weight unaltered during the training stage [20].

**Table 1 Pseudo Code of TVL-1 Algorithm [15]**

```
Input: Two intensity images I₀ and I₁
Output: Flow field u from I₀ to I₁
Pre-process to the input images
For L = 0 to max level do
        Calculate restricted pyramid image ᴸI₀ and ᴸI₁
end
Initialize Lu = 0, ᴸp = 0 and L = maxLevel;
While L >= 0 do
        For W= 0 to maxWraps do
                Re-sample coefficients of p using ᴸI₀, ᴸI₁ and ᴸu
                For Out = 0 to maxOuterIterations do
                        Solve ᴸv using thresholding
                        For In = 0 to maxInnerIterations do
                                Perform each iteration step and solve for ᴸu;
                        End
                        Median-filter Lu;
                End
        End
        If L > 0 then
                Prolongate ᴸu and ᴸp to calculate next pyramid level L -1
        End
End
```

**Note: A Super Scripted L denotes the pyramid level**

## 3.2 Proposed solution

The proposed solution consists of two main stages (see Fig. 2): the pre-processing stage, and the feature extraction and the classification stage.

*Pre-processing stage*

This section describes how the video is represented and feed into the deep CNN as an input image. First, the optical flow is computed from the series of RGB image frames. The calculation for optical flow is crucial because the optical flow captures the object that exists in motion in the video, and thereby it directly captures the dynamic action. The calculation of rank pooling formulation is applied to the generated optical flow images. The rank pooling approach can be applied directly to RGB images, but we have first needed to compute the optical flow and then applied the rank pooling approach to the generated optical flow images. The reason behind this step is that the rank pooling approach captures the motion associated with the action. Besides, it only captures how RGB values change from pixel-to-pixel. The rank pooling approach looks to the increasing of pixel intensities from one frame to others, which are unrelated to action. To mitigate this issue, rank pooling has been applied to optical flow images instead of RGB images. Then, the rank pooling approach used on an optical image, which generates a dynamic optical flow image and then passed it as input to CNN.

*Feature extraction and classification stage:*

This section describes how the features of the dynamic images are extracted and classified using CNN. This stage is adopted from Núñez-Marcos et al. [3]. After getting sufficient temporal evolution from the dynamic images, the dynamic images are fed into the VGG16 Net CNN architecture. The VGG-16 was chosen because of its high accuracy in the feature extraction process compare to other architecture. The VGG16 Net CNN architecture consists of three main layers: convolutional layer, max-pooling layer, and fully connected. These layers are followed by the soft-max function, as shown in Fig. 2. In convolution layers, VGG-16 restricts the use of 3x3 convolution kernels size, which will help to reduce the number of parameters that can result in practical training and testing. Also, using a small kernel size can help to stack more layers in deep networks which would increase the performance of fall detection. CNN takes either an image or feature map as an input while its output is a feature map. The Max-pooling layer makes downsize to the feature map and helps in reducing the over-fitting problem. In VGG architecture, each convolution layer was followed by the rectified linear unit (ReLU) layer. Pooling layers operate on the blocks of the input feature map and combines the featured activations. This combination task is defined by a max function, which selects the maximum activation from the chosen group of blocks. The Max-pooling layer minimizes the spatial size of the feature, and it helps in reducing the number of parameters and the amount of computation in the network [18]. A fully connected layer takes the input from the ReLU layer and generates the class score that is used for the classification of fall detection. VGG16 architecture uses dropout layers in the first two fully connected layers to avoid over-fitting. The input layer of VGG-16 accepts a stack of optical flow images. The ImageNet dataset was utilized to address a small number of fall samples in the dataset. Besides, it used to learn the generic features as it has 14 million images. Based on the CNN-trained ImageNet dataset, the input to the CNN is modified to accept input images of size 224x224 and stack size of 20. Moreover, later fine-tunes to the network to be adopted for the fall detection dataset. The technique of using the knowledge from one dataset and apply the learning features to another dataset is called a transfer learning technique. In the final step of transfer learning, the weight of the convolution layer was frozen to make the weight unaltered during the training stage [18].

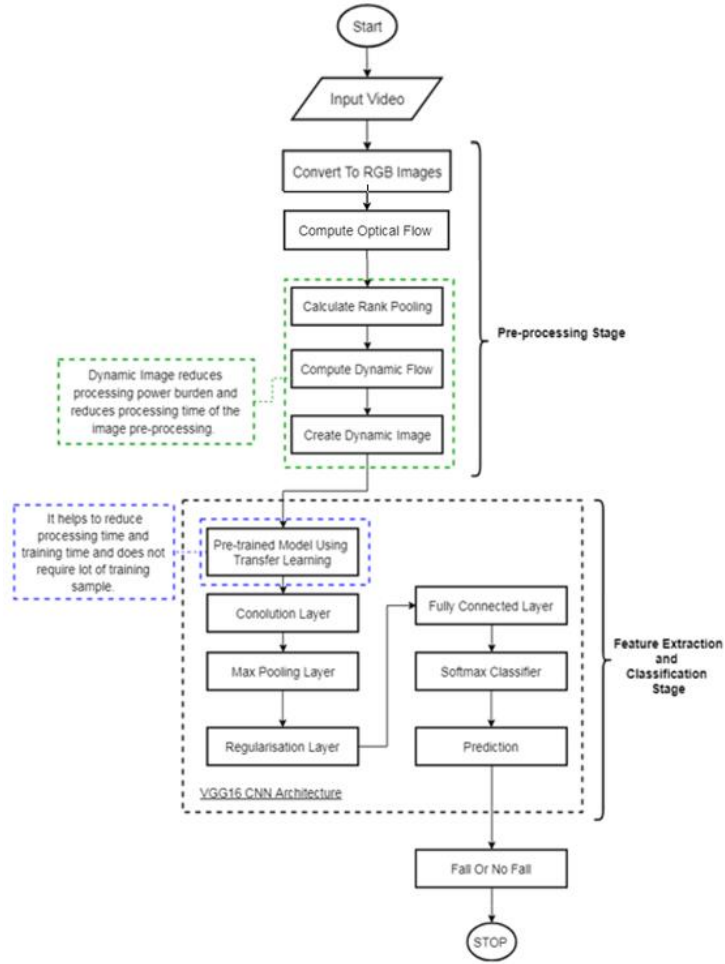

**Figure 2:** Block diagram of the proposed system

[Inherited features of the state-of-the-art solution (in blue) and feature enhanced by proposed solution (in green)]

### 3.3 Proposed equation

A dynamic image is a combination of several consecutive frames in a video, which is based on rank pooling that is capable of capturing the appearance and temporal evolution of the information in the video. According to Bilen et al. [21], a dynamic image can be calculated by using Eq. (2).

$$E = \frac{\lambda}{2}||\rho(I_1,\ldots,I_T;\psi)||^2 + \frac{2}{T(T-1)} \times \sum_{q>t} \max\{0, 1 - S(q|d) + S(t|d)\} \tag{2}$$

Where,
$I_1 \ldots I_T$ all the image frame in the video.
$S(q|d)+S(t|d)$ the score of ranking function at each time t and at later time q.
$\psi(I_t)$ feature vector extracted from each individual frame It in the video.
$\lambda$ is a relative weighting term.
$\rho(I_1,\ldots,I_T;\psi)$ a function that maps a sequence of T video frames to a single vector.
E the energy function.
d is a vector of parameters.

The ranking function is one of the popular methods to represent time series is to apply the atemporal pooling operator to the features extracted at individual time instants [21]. The function parameters d is learned so that the scores reflect the rank of the frames in the video.

Enhanced energy is given by using Eq. (3) :

$$E_e = \sum_{q>t} \max\{0, 1 - S(q|d) + S(t|d)\} \quad (3)$$

Based on the modification that has been made on Eq. (1), this paper introduced a modified energy function (Enhanced Dynamic Optical Flow) as it is illustrated in eq. (3). We have added a ranking function from Eq. (4) to the first term of Eq. (1). This will be added temporal dynamics of the action in the image frames ($I_1$ and $I_2$) and solve the accuracy problem in dynamics lighting condition.

$$E_m = \int_\Omega \{\lambda|I_1(x) - I_2(x+u(x))\}dx + E_e \quad (4)$$

### 3.4 Area of improvement

This paper proposed an enhanced optical dynamic flow, which is the extent of the optical flow and rank pooling approach. With the help of Eq. (4), we provide the extension of optical flow, and dynamic images that help in capturing the dynamics fall action directly while the threshold of the flow vectors helps in avoiding the background noise. The temporal pooling formulation has been applied to the flow image instead of just RGB images to capture the long-term temporal context directly. Additionally, the difficulties have been represented in achieving long-term dependency in optical flow algorithms. Temporal pooling was used to generate optical flow images and dynamic flow images, which then fed them into CNN as an input. The generation for dynamic optical flow images is very efficient: extracting features from them is fast and straightforward, and reducing the analysis of full videos to analyze a single image significantly accelerates recognition.

### 3.5 Why enhanced optical dynamic flow (EODF)?

In the proposed solution, the main benefit of using a dynamic flow image is to summarize the whole video, which can capture the long-term temporal extent of the human action (fall action) of a video in just one image. Thus, reducing the processing time of analyzing the video to a single image. Since individual fall event consists of more than just basic human posture such as sitting, lying, standing, and falling, it is inefficient to use a short clip consisting of only a few frames. The result leads to an increase in the accuracy of fall detection in complex environments such as dynamic lighting conditions. The current state-of-the-art approach has not been regarded as the temporal evolution of the action. By proposing optical flow techniques, only a few image frames have been used and passed as input to the CNN.

## 4. Results and discussion

In this work, three different datasets are used: UR fall dataset (URFD), multiple cameras (Multicam), and Fall detection dataset (FDD). FDD consists of 191 videos that are annotated for evaluation purposes. It has additional information representing the ground truth of the fall position in the image sequence. Multicam fall dataset consists of 24 scenarios recorded with eight different video cameras. URFD dataset consists of 70 video sequences (30 videos of falls and 40 videos of activities of daily living). All the fall events are recorded within 2 Microsoft Kinect cameras. The tested videos of the utilized datasets have a resolution of 320x240 with 25fps from different environments. These three datasets are recorded in a controlled environment with certain restrictions to get a better result. First, there is just one actor in all the used videos. Besides, the videos are recorded under suitable lighting conditions. Moreover, each video contains a frame number of the beginning of the fall, frame number of the end of the fall, the height, the width, and the coordinate of the center of the bounding box for each frame. However, to test the proposed model dynamic lighting environment, dynamic artificial lighting was added for each frame, channel, and pixel. A progressive change of lighting was added to the dynamic artificial lighting to increase its intensity from frame to frame until reaching the desired value. This lighting modification was added once per video at its first 32 frames. A single lighting change was used for each video to add more realistic lighting conditions.

The large samples include different training epochs, batch size, and symbols were selected from the three datasets. Two standard datasets are the fall detection dataset (FDD) and URFD used for evaluation. These datasets are open source and freely available on the internet as a .mpeg file. These .mpeg are pre-processed to create a series of images via optical flow algorithm using Python language. Fall detection can be thought of as a binary classification problem according to the occurrence of the fall in the used video. That is, whether a video contains a fall or not. The standard metrics used to assess the performance of the classier are sensitivity (also known as a true positive rate or recall), and specificity(also known as a true negative rate). An imbalance of class distribution does not bias these metrics. It makes them suitable for the fall detection dataset as the number of samples in the fall detection dataset is lower than other datasets. During the pre-processing stage of the proposed system, the TVL1 optical flow algorithm converts each video from the dataset (URFD and FDD) into optical flow images. After that, these images are passed as input to CNN to train the network. CNN performs the training on the split datasets. Once the training is completed, the model is saved in a .hd file to be used for further process. In the testing stage, the saved HD file is loaded into the network.

During testing, the network is fine-tuned to get the feature for correct classification actions. To get the ultimate training data, the raw video training data should be trimmed into segments according to the number of phase classifications. Then each of these video clips should be converted to a dynamic image and used as the data for fine-tuning the network parameters. This was accomplished by first replicating the weights from the pre-trained VGG-16 network, except for the last fully connected layer and the softmax classifier. The last fully-connected layer of the pre-trained CNN should be replaced with a new layer that has four neurons for the four-class classification task. The last fully connected layer is randomly initiated. The last layer is tune only, followed by tuning all layers in the CCN. During testing, the network is fine-tuned to get the feature for correctly classification actions, which is mean the fun-tuning procedure is used for the feature extraction process. Based on the training and testing of the data using the proposed system, accuracy and loss result are predicted, as shown in Fig. 3, Fig. 4, and Table 2.

Table 2. Accuracy and Processing Time Results for Proposed and state of the art solutions

| Sample details From different datasets | Original image | State of the Art [3] | | | Proposed solution | | |
|---|---|---|---|---|---|---|---|
| | | Processed sample/output | Accuracy by probability score in classification | Processing time (second) | Processed sample | Accuracy by probability score | Processing time |
| URFD Dataset | Sample: fall-27, Camera 0, RGB Data | | | | | | |
| | 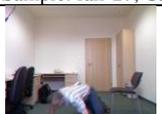 | 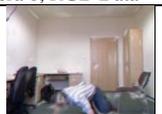 | 94.58% | 0.480s | 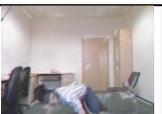 | 96.41% | 0.398s |
| | Sample: fall-30, Camera 0, RGB Data | | | | | | |
| | 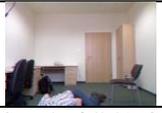 | 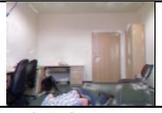 | 97.23% | 0.472s | 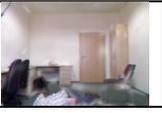 | 93.23% | 0.401s |
| | Sample: fall-29, Camera 0, RGB Data | | | | | | |
| | 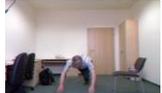 | 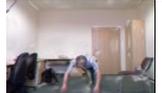 | 88.22% | 0.412s | 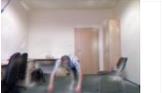 | 92.45% | 0.385s |
| | Sample: fall-16, Camera 0, RGB Data | | | | | | |
| | 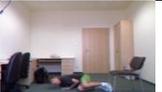 | 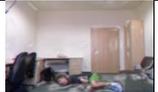 | 91.34% | 0.434s | 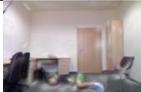 | 95.74% | 0.398s |
| | Sample: fall-10, Camera 0, RGB Data | | | | | | |
| | 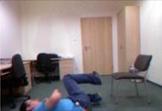 | 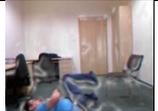 | 97.41% | 0.471s | 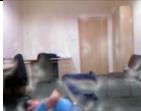 | 98.14% | 0.414s |
| | Sample: fall-01, Camera 0, RGB Data | | | | | | |
| | 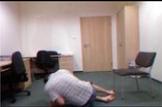 | 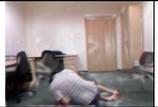 | 93.98% | 0.465s | 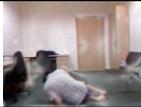 | 94.72% | 0.475s |
| | Sample: fall-12, Camera 0, RGB Data | | | | | | |
| | 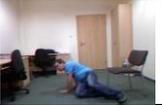 | 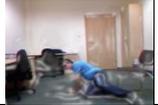 | 92.74% | 0.482s | 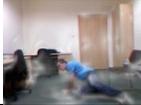 | 95.08% | 0.387s |
| Multiple cameras fall dataset | Sample: Scenario 01, Camera 07, Multiple cameras fall dataset | | | | | | |
| | 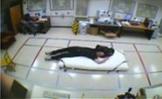 | 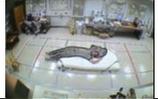 | 92.51% | 0.471s | 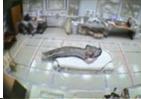 | 93.86% | 0.396s |
| | Sample: Scenario 01, Camera 01, Multiple cameras fall dataset | | | | | | |
| | 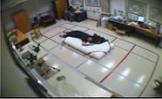 | 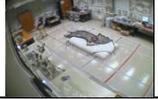 | 94.45% | 0.425s | 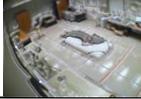 | 95.63% | 0.357s |
| | Sample: Scenario 01, Camera 01, Multiple cameras fall dataset | | | | | | |
| | 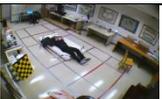 | 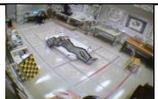 | 92.84% | 0.429s | 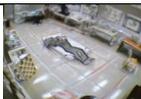 | 94.15% | 0.375s |
| | Sample: Scenario 01, Camera 05, Multiple cameras fall dataset | | | | | | |
| | 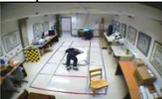 | 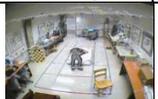 | 89.30% | 0.439s | 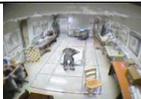 | 92.23% | 0.381s |
| | Sample: Scenario 01, Camera 08, Multiple cameras fall dataset | | | | | | |
| | 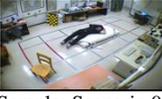 | 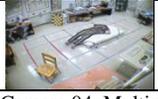 | 91.09% | 0.427s | 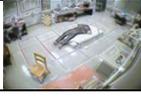 | 93.36% | 0.395s |
| | Sample: Scenario 01, Camera 04, Multiple cameras fall dataset | | | | | | |
| | 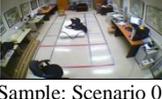 | 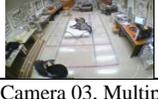 | 87.25% | 0.431s | 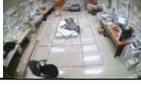 | 89.4% | 0.401s |
| | Sample: Scenario 01, Camera 03, Multiple cameras fall dataset | | | | | | |
| | 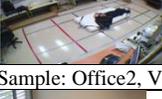 | 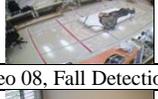 | 90.29% | 0.412s | 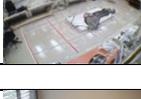 | 91.75% | 0.385s |
| Fall Detection Dataset | Sample: Office2, Video 08, Fall Detection Dataset | | | | | | |
| | 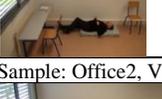 | 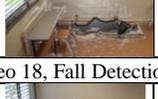 | 91.02% | 0.472s | 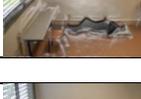 | 95.77% | 0.417s |
| | Sample: Office2, Video 18, Fall Detection Dataset | | | | | | |
| | 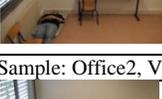 | 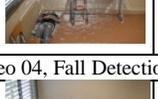 | 91.47% | 0.478s | 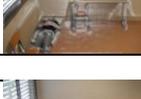 | 94.91% | 0.389s |
| | Sample: Office2, Video 04, Fall Detection Dataset | | | | | | |
| | 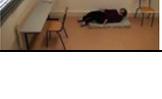 | 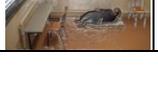 | 88.74% | 0.482s | 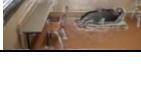 | 91.23% | 0.401s |

| | Sample | | | | | | |
|---|---|---|---|---|---|---|---|
| | Sample: Office2, Video 15, Fall Detection Dataset | | | | | | |
| | 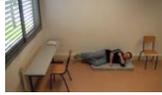 | 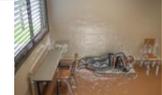 | 90.44% | 0.486s | 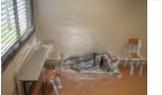 | 96.57% | 0.411s |
| | Sample: Office2, Video 22, Fall Detection Dataset | | | | | | |
| | 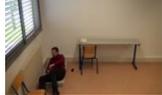 | 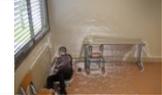 | 87.25% | 0.492s | 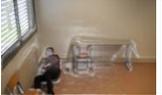 | 93.21% | 0.392s |
| | Sample: Office2, Video 19, Fall Detection Dataset | | | | | | |
| | 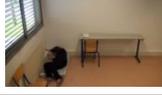 | 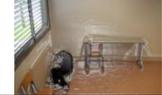 | 89.28% | 0.472s | 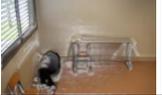 | 93.92% | 0.385s |
| | Sample: Office2, Video 09, Fall Detection Dataset | | | | | | |
| | 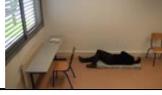 | 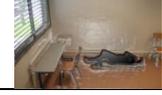 | 92.72% | 0.447s | 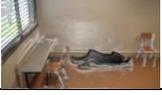 | 95.27% | 0.382s |
| Fall Detection Dataset (Dark Lighting Condition) | Sample: Home1, Video 03 (Brightness: -40%), Fall Detection Dataset | | | | | | |
| | 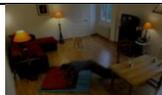 | 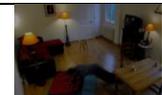 | 87.24% | 0.593s | 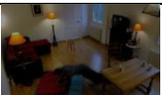 | 89.47% | 0.486s |
| | Sample: Home1, Video 07(Brightness: -40%), Fall Detection Dataset | | | | | | |
| | 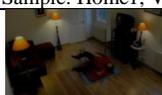 | 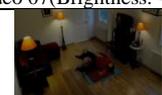 | 86.15% | 0.625s | 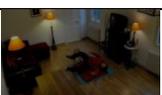 | 87.25% | 0.536s |
| | Sample: Home1, Video 10 (Brightness: -40%), Fall Detection Dataset | | | | | | |
| | 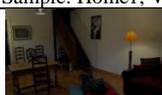 | 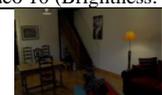 | 83.3% | 0.591s | 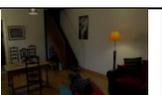 | 83.94% | 0.512s |
| | Sample: Home1, Video 15 (Brightness: -40%), Fall Detection Dataset | | | | | | |
| | 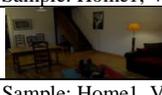 | 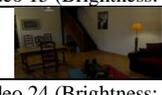 | 84.07% | 0.529s | 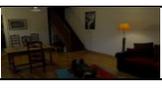 | 85.85% | 0.481s |
| | Sample: Home1, Video 24 (Brightness: -40%), Fall Detection Dataset | | | | | | |
| | 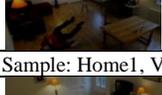 | 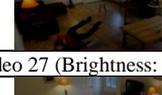 | 84.88% | 0.582s | 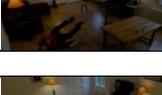 | 86.18% | 0.495s |
| | Sample: Home1, Video 27 (Brightness: -40%), Fall Detection Dataset | | | | | | |
| | 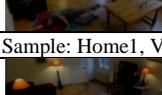 | 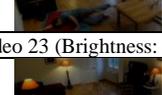 | 83.27% | 0.674s | 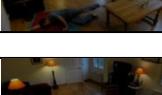 | 84.36% | 0.574s |
| | Sample: Home1, Video 23 (Brightness: -40%), Fall Detection Dataset | | | | | | |
| | 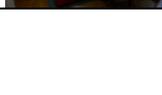 | 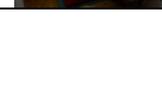 | 82.28%S | 0.685s | 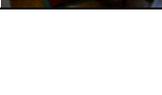 | 85.47% | 0.602s |

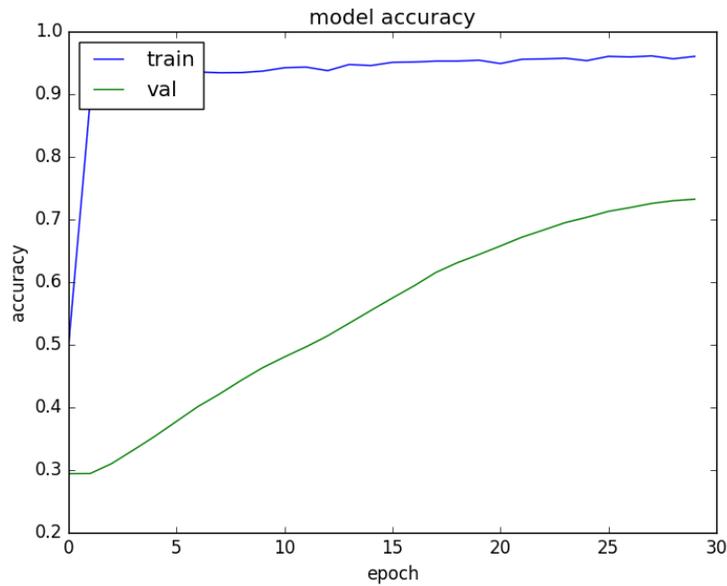

Figure 3: Predicted accuracy of classifier (number of epoch on the x-axis and accuracy level is on the y-axis)

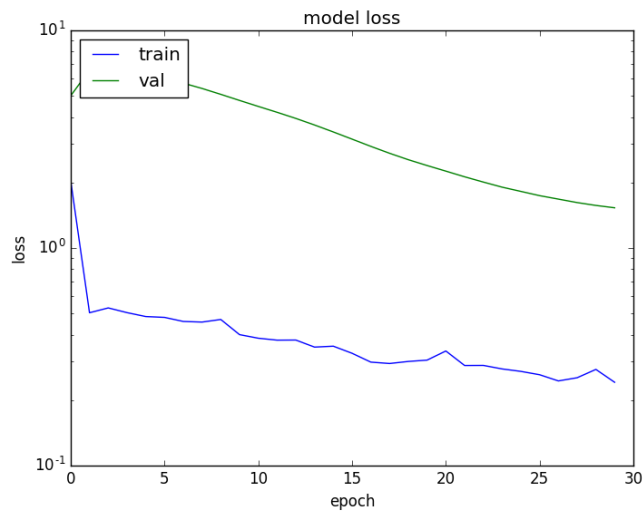

Figure 4: Predicted loss of classifier (number of epoch on the x-axis and loss level is on the y-axis)

Tables and graphs were used to make a comparison between the state-of-the-art and the proposed solution. Samples of frames were compared between the state-of-the-art and the proposed solutions for the fall detection problem. The results from the fall sequence samples are reviewed in Table 2. All samples in the tables contain the results obtained during the classification of human fall sequences in various scenarios. The results from the sample are presented in terms of accuracy, precision and Recall, and processing time. Accuracy is calculated in terms of the specificity and sensitivity of the classifier. A comprehensive test for five samples and three scenarios for each sample was performed. The system accuracy was calculated by taking the average results of each test case in the input samples. Then, the final result is calculated by taking the average for all test cases in the three scenarios.

These results were compared during different stages of image pre-processing and classification of deep learning in different environments with varying lighting conditions. The obtained results showed that the proposed solution improved the accuracy of fall classification in dynamic lighting conditions. Also, the proposed solution reduced the processing time of image pre-processing by increasing the number of frames that processed per second. The proposed system used open-source datasets (FDD and URFD), and it can be applied to other standard fall datasets as well. See figures 5 and 6.

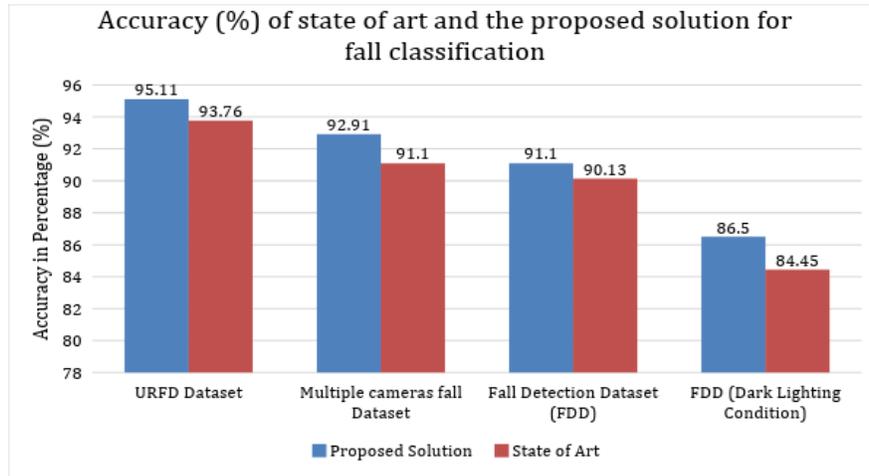

Figure5: Comparison of accuracy (%) of state-of-the-art and the proposed solution for fall classification

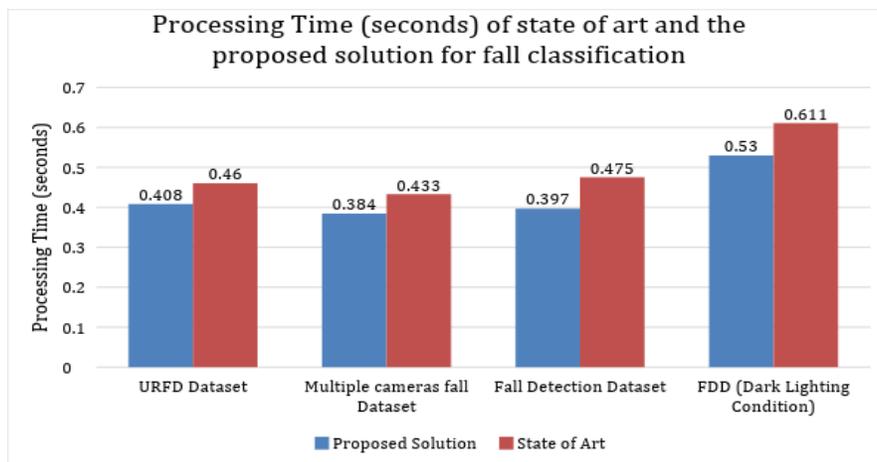

Figure 6: Comparison of processing time of state-of-the-art and the proposed solution for fall classification

The experimental results showed a significant difference in the overall performance of the state-of-the-art and proposed solution. The proposed solution improves the processing time of the system by 40 to 50ms. This result was achieved by using an FC-NN binary classifier and dynamic optical flow in the pre-processing stage. Three standard datasets have been used for training and testing. The system obtained accuracy in terms of specificity and sensitivity around 93.04% overall for the entire environment. The accuracy is calculated based on the probability score assigned for each class using FC-NN binary classifier. The processing time of the proposed model and the state-of-the-art model was calculated in the same way. The processing time was calculated by finding the execution time of both models and finding their overall differences. We quantified processing time in terms of seconds. It was calculated by finding the difference between the beginning time of falling and the ending time of the predicted temporal extent of the falling phase. The total enhancement in system processing time was quantified with 0.396 seconds. The dynamic images improve the performance of the images and help to reduce the processing burden in the pre-processing stage. The dynamic image is created by summarizing a video clip that has shown a falling event into a single image. Fan et al. [1] solved the issue of temporal evolution in videos by using dynamic images. The use of dynamic images for pre-processing a video into an image not only reduces the processing time but also enhances the performance of the fall detection classifier in various lighting conditions. In conclusion, the combination of dynamic images and optical flow techniques improves the performance of fall detection classification and reduces the processing time of the pre-processing stage. See Table 3.

**Table 3. Comparison between state of the art and the proposed solution.**

|  | **Proposed Solution** | **State of the art Solution** |
| --- | --- | --- |
| **Name of the solution** | Enhanced Dynamic Optical Flow Algorithm | TVL-1 Optical Flow Algorithm |
| **Accuracy** | Improved accuracy in terms of sensitivity of fall detection in dynamic lighting conditions. | Improved accuracy in terms of sensitivity of fall detection in stable lighting conditions. |
| **Processing Time** | Reduced processing time of the image pre-processing stage using enhanced dynamic optical flow | Processing time and required processing powers of the pre-processing stage are higher. |
| **Contribution 1** | Dynamic flow image to summarize the whole video is to capture long term temporal extent of the human action | The current state-of-the-art approach has not considered the temporal evolution of the action. |
| **Contribution 2** | Dynamic flow image to reduce the processing power required to pre-process the video. | The current state-of-the-art approach requires a lot of processing power to process the video into an image sequence. |

## 5. Conclusion:

This research aims to propose a vision-based fall detection system that improves the accuracy of fall detection in some complex environments such as the change of light condition in the room. The proposed system concentrates on decreasing the processing time of fall detection and improving classification accuracy. Meanwhile, it provides a mechanism for summarizing a video into a single image by using a dynamic optical flow technique, which helps to increase the performance of image pre-processing steps. The proposed solution has achieved high sensitivity in terms of classification of fall detection in dynamic lighting conditions. The solution applied dynamic optical flow technique and deep learning in the image pre-processing stage. The proposed solution solves the limitations in the pre-processing stage of the state-of-the-art system. The current techniques have not considered the long-term temporal evolution of an action, which is crucial activates like fall events. Our solution overcomes the issue in the TVL-1 optical flow technique by using a rank pooling approach and dynamic image. The accuracy of the system has been improved by around 3% and the processing time by 40 to 50ms.

Sagar Chhetri, Abeer Alsadoon, Thair Al-Dala'in, P. W. C. Prasad, Tarik A. Rashid, Angelika Maag (2020). Deep learning for vision-based fall detection system: Enhanced optical dynamic flow. Computational Intelligence. DOI:https://doi.org/10.1111/coin.12428## References:

[1] Fan, Y., Levine, M., Wen, G., & Qiu, S. (2017). A deep neural network for real-time detection of falling humans in naturally occurring scenes. *Neurocomputing*, 260, 43-58. doi: 10.1016/j.neucom.2017.02.082

[2] Wang, S., Chen, L., Zhou, Z., Sun, ..X., & Dong, J. (2016). Human fall detection in surveillance video based on PCANet. *Multimedia Tools And Applications*, 75(19), 11603-11613. doi: 10.1007/s11042-015-2698-

[3] Núñez-Marcos, A., Azkune, G., & Arganda-Carreras, I. (2017). Vision-Based Fall Detection with Convolutional Neural Networks. *Wireless Communications And Mobile Computing*, 2017, 1-16. doi: 10.1155/2017/9474806

[4] H. Sadreazami, M. Bolic and S. Rajan, "Fall Detection Using Standoff Radar-Based Sensing and Deep Convolutional Neural Network," in *IEEE Transactions on Circuits and Systems II: Express Briefs*, vol. 67, no. 1, pp. 197-201, Jan. 2020, doi: 10.1109/TCSII.2019.2904498

[5] Zerrouki, N., Harrou, F., Sun, Y., & Houacine, A. (2018). Vision-Based Human Action Classification Using Adaptive Boosting Algorithm. *IEEE Sensors Journal*, 18(12), 5115-5121. doi: 10.1109/jsen.2018.2830743

[6] M. Saleh and R. L. B. Jeannès, "Elderly Fall Detection Using Wearable Sensors: A Low Cost Highly Accurate Algorithm," in *IEEE Sensors Journal*, vol. 19, no. 8, pp. 3156-3164, 15 April15, 2019, doi: 10.1109/JSEN.2019.2891128.

[7] Ali, S., Khan, R., Mahmood, A., Hassan, M., & Jeon, a. (2018). Using Temporal Covariance of Motion and Geometric Features via Boosting for Human Fall Detection. *Sensors*, 18(6), 1918. doi: 10.3390/s1806191

[8] Xiong, X., Min, W., Zheng, W. *et al.* S3D-CNN: skeleton-based 3D consecutive-low-pooling neural network for fall detection. *Appl Intell* (2020). https://doi-org.ezproxy.uws.edu.au/10.1007/s10489-020-01751-y

[9] Min, W., Cui, H., Rao, H., Li, Z., & Yao, L. (2018). Detection of the human Falls on Furniture Using Scene Analysis Based on Deep Learning and Activity Characteristics. *IEEE Access*, 6, 9324-9335. doi: 10.1109/access.2018.2795239

[10] Zhang ZM, Ma X, Wu HB, Li YB (2019) Fall detection in videos with trajectory-weighted deep-convolutional rank-pooling descriptor. IEEE Access 7:4135–4144. https://doi-org.ezproxy.uws.edu.au/10.1109/Access.2018.2887144

[11] Zerrouki, N., & Houacine, A. (2017). Combined curvelets and hidden Markov models for human fall detection. *Multimedia Tools And Applications,* 77(5), 6405-6424. doi: 10.1007/s11042-017-4549-5

[12] L. Wang, M. Peng and Q. Zhou, "Pre-Impact Fall Detection Based on Multisource CNN Ensemble," in *IEEE Sensors Journal*, vol. 20, no. 10, pp. 5442-5451, 15 May15, 2020, doi: 10.1109/JSEN.2020.2970452.

[13] W.-Cheng and D.-M. Jhan, "Triaxial accelerometer-based fall detection method using a self-constructing cascade-AdaBoost-SVM classifier", *IEEE J. Biomed. Health Inform.*, vol. 17, no. 2, pp. 411-419, Mar. 2013.

[14] N. Hnoohom, A. Jitpattanakul, P. Inluergsri, P. Wongbudsri and W. Ployput, "Multi-sensor-based fall detection and activity daily living classification by using ensemble learning", *Proc. Int. ECTI Northern Sect. Conf. Electr. Electron. Comput. Telecommun. Eng. (ECTI-NCON)*, pp. 111-115, Feb. 2018.

[15]Sánchez Pérez, Javier, Meinhardt-Llopis, Enric, & Facciolo, Gabriele. (2013). TV-L1 Optical Flow Estimation. *Image Processing on Line, 3*, 137-150.

[16] Wedel, Andreas & Pock, Thomas & Zach, Christopher & Bischof, Horst & Cremers, Daniel. (2009). An Improved Algorithm for TV-L1 Optical Flow. 10.1007/978-3-642-03061-1_2.

[17] Pingault, M & Pellerin, Denis. (2002). Optical flow constraint equation extended to transparency.

[18] Hamprecht, F. A., Schnörr, C. Jähne, B (2007). A duality based approach for realtime TV-L 1 optical flow. *Pattern Recognition*. pp. 214–223.

[19] Wedel A., Pock T., Zach C., Bischof H., Cremers D. (2009). An Improved Algorithm for TV-L1 Optical Flow. Statistical and Geometrical Approaches to Visual Motion Analysis. Springer

[20] Khan, S., Rahmani, H., Shah, S. A. A., & Bennamoun, M. (2018). A Guide to Convolutional Neural Networks for Computer Vision. (Synthesis Lectures on Computer Vision; Vol. 8, No. 1). Morgan & Claypool Publishers. DOI: 10.2200/S00822ED1V01Y201712COV015

[21] Bilen, H., Fernando, B., Gavves, E., & Vedaldi, A. (2017). Action Recognition with Dynamic Image Networks. *IEEE Transactions On Pattern Analysis And Machine Intelligence*, 1-1. doi: 10.1109/tpami.2017.2769085.

16